# End-User Construction of Influence Diagrams for Bayesian Statistics

Harold P. Lehmann, MD PhD
Johns Hopkins University
Baltimore, MD 21205
lehmann@welchgate.welch.jhu.edu

Ross D. Shachter, PhD
Stanford University
Stanford, CA 94305
shachter@camis.stanford.edu

## Abstract

Influence diagrams are ideal knowledge representations for Bayesian statistical models. However, these diagrams are difficult for end users to interpret and to manipulate. We present a user-based architecture that enables end users to create and to manipulate the knowledge representation. We use the problem of physicians' interpretation of two-arm parallel randomized clinical trials (TAPRCT) to illustrate the architecture and its use. There are three primary data structures. Elements of statistical models are encoded as subgraphs of a restricted class of *influence diagram*. The interpretations of those elements are mapped into users' language in a domain-specific, user-based semantic interface, called a *patient-flow diagram*, in the TAPRCT problem. Permitted transformations of the statistical model that maintain the semantic relationships of the model are encoded in a *metadata-state diagram*, called the cohort-state diagram, in the TAPRCT problem. The algorithm that runs the system uses modular actions called *construction steps*. This framework has been implemented in a system called THOMAS, that allows physicians to interpret the data reported from a TAPRCT. Keywords: statistics, Bayesian, expert systems, artificial intelligence, influence diagrams, probabilistic reasoning, knowledge acquisition, clinical trials.

## 1 INTRODUCTION

Dynamic construction of influence diagrams is a powerful strategy for enabling a computer-based system to match a human's knowledge. During statistical analysis, the analyst creates a model of the domain and of the data, so such dynamic construction is crucial when creating intelligent systems for statistical analysis. Influence diagrams are ideal knowledge representations for statistical models, especially for Bayesian statistical models (Smith 1989). However, the use of Bayesian statistics needs the end user to direct much of the analysis, because of the importance of the user's prior knowledge of the domain. Unfortunately, by the very nature of the complexity and technicality of Bayesian statistical models, the typical end structures of these models or the implications of any choices for important issues like methodological concerns or prior knowledge.

Our thesis is that a user-based semantic interface to influence-diagram construction will allow such users to perform a meaningful Bayesian-statistical analysis. We shall demonstrate a program architecture we call user based (UB) that can serve as the basis for building a wide variety of such systems. We shall target a particular user community throughout this paper: physician readers of reports of clinical trials, specifically, two-arm parallel ran-domized clinical trials (TAPRCT). These are studies where patients are randomly assigned to one of two therapies; often, one of them is a placebo control treatment. The principles of using the architecture, however, are generic.

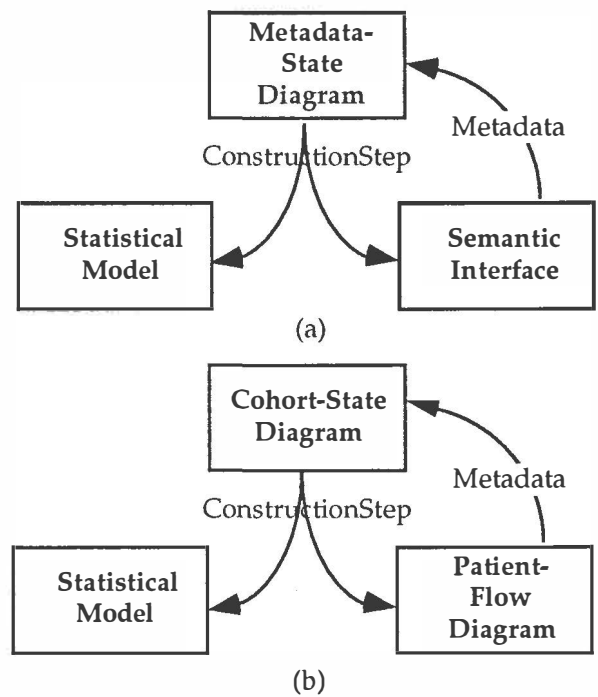

Figure 1: The System architecture. (a) General. (b) Particular to the case of the TAPRCT problem.

Table 1: Elements of the influence-diagram representation of statistical models.

| Object Class | Influence-Diagram Element | Statistical-Model Interpretation |
|---|---|---|
| Level | Population | General knowledge |
|  | Sample | Study-specific knowledge |
|  | Effective | Measured knowledge |
|  | Patient | Patient-specific knowledge |
| Node | Outcome | e.g., *Lifespan* |
|  | Parameter |  |
|  |   Outcome | e.g., *Mortality rate* |
|  |   Methodological | e.g., *Withdrawal rate* |
| Arc | Interlevel |  |
|  |   Population→ Study | External validity e.g., *Selection bias* |
|  |   Study → Effective | Measurement reliability e.g., *Specificity* |
|  | Isotypal |  |
|  |   Population → population | Domain knowledge |
|  |  | e.g., *Placebo population mortality rate = baseline population mortality rate* |
|  |   Sample → sample | Internal validity e.g., *Withdrawal model* |
|  | Heterotypal |  |
|  |   Methodological → outcome | Specific relationships e.g., *Mixture models* |

Fig. 1 depicts the architecture. In Section 2, we shall discuss statistical models, in Section 3, the semantic interface, in Section 4, metadata and the metadata-state diagram, and in Section 5, the construction steps. In Section 6, we shall discuss an implementation of this architecture, in Section 7 we discuss previous work, and conclude in Section 8.

## 2 STATISTICAL MODEL

A statistical model comprises *elements, interpretations*, and *transformations*. Elements of a statistical model include: probability models, parameters, observations, and assumptions that allow the analyst to relate the observations to the parameters. Interpretations map elements of the statistical model into inferential concerns in the world. Transformations are the permitted operations that change the statistical model, while maintaining the interpretations of the model.

In the UB architecture, we use *influence diagrams* (Howard and Matheson 1981; Oliver and Smith 1990) to represent the elements of statistical models (Smith 1989). As generally described, there are two classes of objects in these acyclic, directed graphs. *Nodes* are the objects of interest: parameters, observations, decision options and decision objectives. Note that parameters and observations are continuous quantities, unlike the discrete variables for which influence diagrams have been traditionally used in AI systems. *Arcs* are objects representing dependencies; the absence of an arc between two nodes denotes the independence of the variables those nodes represent.

To aid the communication with users, we add *level* as a third object. Levels are elements that contain nodes and give those nodes interpretive semantics (e.g., *population* mortality rate as opposed to *sample* mortality rate). Levels limit the transformative interactions among parameters. The specific levels we use for the TAPRCT problem are as follows (see Table 1): *effective* level allows for modeling concerns regarding measurement reliability; the *sample* level allows for modeling concerns regarding internal validity; the *population* level allows for modeling concerns regarding external validity; and the *patient* level allows us to represent differences between the patient and the population. Fig. 2 depicts the initial influence-diagram of the statistical model for the TAPRCT problem.

In the UB architecture, we are concerned primarily with transformations that signify methodological concerns noted by the reader of the study report. We use *construction steps* that transform the statistical model to comply with the interpretative meaning of the methodological concern (see Section 5). The steps are modular statements of permitted transformations. This modularity will allow the user to take as much charge of the interaction with the machine as possible. Note that modification of a statistical model is, by definition, *model driven*; the nature of the model permits or forbids different types of modification.

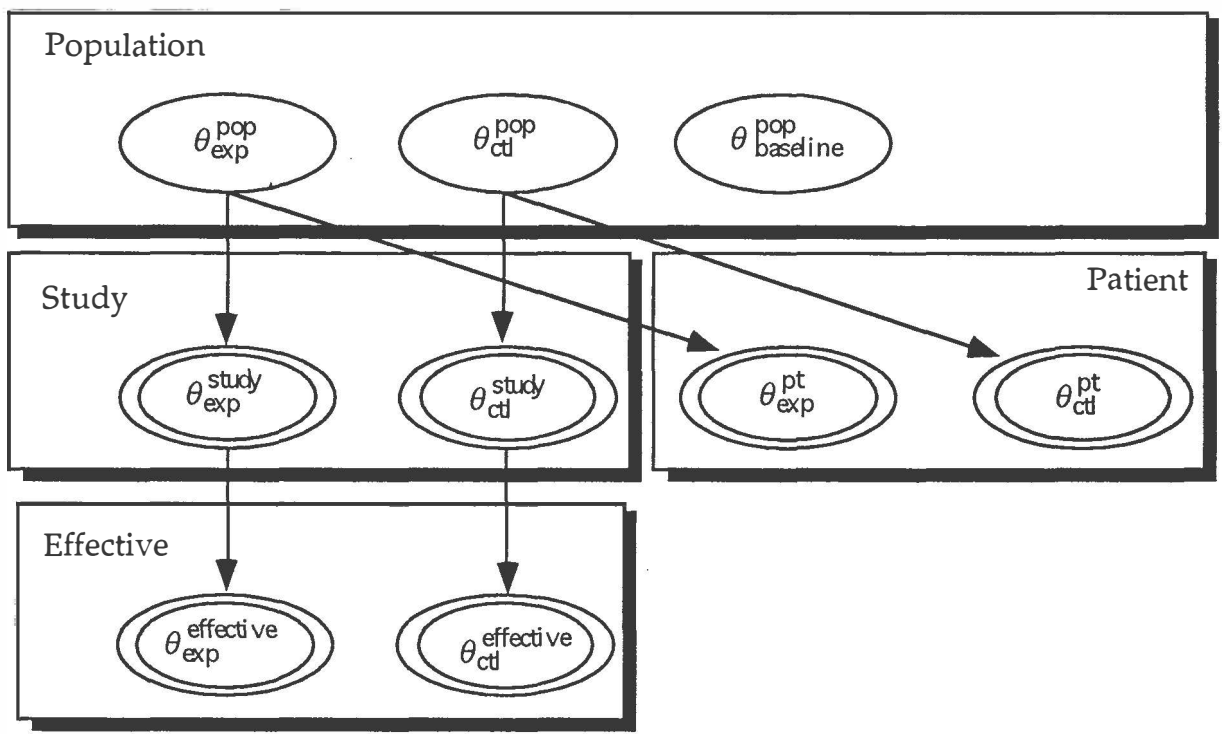

Figure 2: Initial influence-diagram of the statistical model for a TAPRCT. The diagram is divided into four levels (*Population, Study, Effective, and Patient*), with experimental (*exp*) and control (*ctl*) outcome parameters ($\theta$) in each level (plus a *baseline* outcome parameter in the population level). The parameters in the study, effective, and patient levels are functionally dependent on the population-level parameters; the latter are represented as oval chance nodes, the former, as doubly bordered deterministic nodes.

Fig. 3 exhibits the representation of a particular bias in an TAPRCT, showing a part of the influence-diagram representation of a trial before and after the construction step of including the methodological concern of withdrawal[1] (for simplicity, only the study level is shown). The construction has remodeled the dependence of the study parameter: no longer identical to a population parameter (the initial, default relationship), it is now a function of a mixture of other parameters.[2] The mixture parameter represents the withdrawal rate. In a classical-statistical system, its value would be estimated from the data in the study. In a Bayesian system, its value would depend on the prior belief of the analyst—what withdrawal rate she would expect in this type of study—as well as on the data observed. The modular concept for this bias is the recognition that withdrawals affects the study parameter only, and in a particular way.

## 3  SEMANTIC INTERFACE: THE PATIENT-FLOW DIAGRAM

Most statistical clients require an interface that translates elements of a statistical model into a language whose semantics they understand. For the UB architecture, we chose as the metaphor for the semantic interface the *patient-flow diagram*. In such a diagram, children nodes represent cohort at a particular point in time. Modification of a cohort is akin to a traversal from one state of knowledge to cohorts of patients whose numbers sum to that of the parent node, but each of whom experienced an important different event in the study. A patient-flow diagram depicts the history of patients in the course of a study and is often published in a report of a clinical trial. Users manipulate the statistical model by manipulating the patient-flow diagram—the domain-based metaphor. For instance, they might select a cohort and communicate (e.g., via a pop-up menu) that some patients in the cohort withdrew from the assigned therapy. Note that, in contrast to the statistical model, modification of the patient-flow diagram is *data* driven.

---

[1] *Withdrawal* refers to patients who no longer follow the study protocol, but about whom outcome data are known.

[2] Left out of Fig. 2 are the dependencies of the other study parameters on particular population parameters. For instance, the outcome study parameter in patients assigned to experimental treatment but who withdrew from therapy is made functionally identical on the standard-care population parameter.

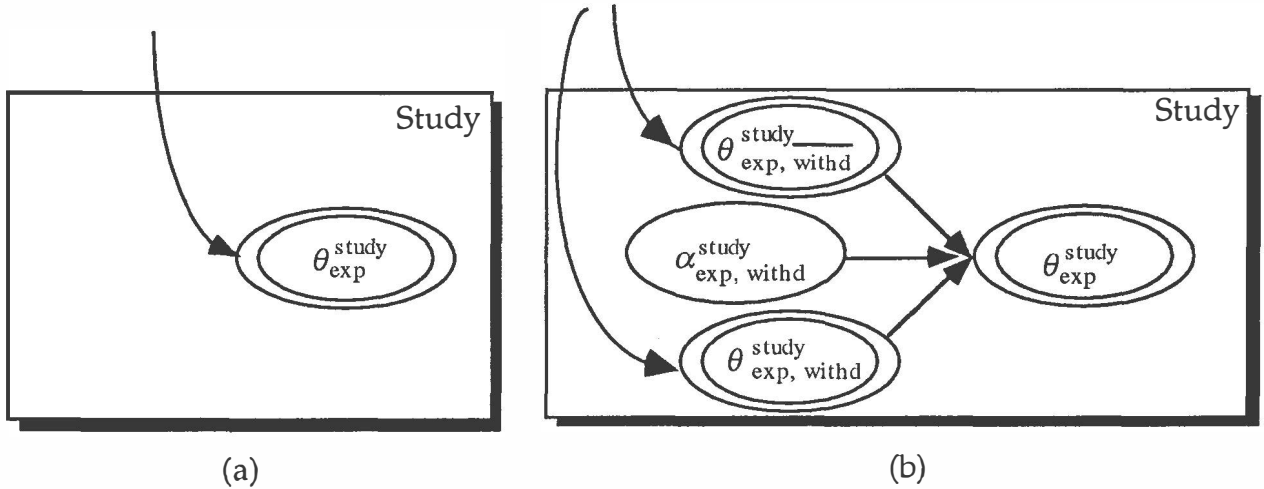

(a)    (b)

Figure 3: Influence-diagram representation of adding the potential bias of withdrawals. (a) Before adding the bias. Only the study-level experimental outcome parameter ($\theta$) shown. The dependency of the parameter on a population-level parameter is indicated by the incoming arc. (b) After adding the bias. The study-level outcome parameter is now dependent on two other outcome parameters, one for each subcohort (the subcohort of patients who did withdraw from the study (exp,withd) and the subcohort where they did not), as well as a new, methodological parameter; this a parameter is called a mixture parameter. The study-level dependence on the population level is now through the subcohort outcome parameters

## 4 METADATA-STATE DIAGRAM: THE COHORT-STATE DIAGRAM

The metadata-state diagram (MDSD) mediates between the user's data-driven manipulation of the patient-flow term refers to knowledge about what sorts of information are contained in cohorts and in nodes, and what modifications are permitted. The MDSD for the TAPRCT problem is called the cohort-state diagram. (see Fig. 4). The state in this diagram refers to the state of knowledge regarding a another; implementing that change of state results in changing the statistical model, as well as the patient-flow diagram. The diagram aspect of the MDSD is the set of path traversals permitted. For the TAPRCT problem, it represents, for instance, the methodological knowledge (based on the definition of loss to followup) that if a statistical model currently reflects the fact that a cohort of patients has been lost to followup, then the user's attempt to provide evidence about that cohort should be denied; the traversal form state $A$ to $B$ in Fig. 4 is forbidden.

In summary, a cohort in the patient-flow diagram points to a state in the metadata-state diagram. The cohort also points to study and effective outcome parameters (such as mortality rates) in the influence diagram (statistical model). Evidence within a cohort becomes a node in the influence diagram, dependent on the effective outcome parameter associated with the cohort. The relationship between population outcome parameter and study outcome parameter reflects domain knowledge.

diagram and the program model-driven modification of the statistical model. Metadata refers to information about data (Chytil 1986). In the UB architecture, this

## 5 CONSTRUCTION STEPS

The top-level controlling algorithm works as follows: The user completes an interaction with the patient-flow diagram. The patient-flow diagram then translates the user's metadata directive into a machine-usable format that includes the metadatum and the target cohort. Unless the directive signals termination of the modeling process, the system examines the cohort-state diagram to determine whether the directive is permitted, by inspecting the arcs emanating from the target cohort's state in the diagram. If the directive is permitted, the system then executes the corresponding construction step, modifying the patient-flow diagram and the statistical model in the process. If new statistical parentless parameters are created by the construction step, the user is asked to assess prior beliefs about those parameters; the constructed names of the parameters are meaningful to the physician user.

Names for new parameters are created by concatenating phrases associated with transitions in the metadata-state diagram, modified by knowledge available to the system regarding the target cohort. For instance, a new withdrawal-cohort study outcome parameter will be called *study \<outcome\> \for patients who withdrew from therapy in \<name of parental cohort\>*, where \<outcome\> might be *mortality*, \

might be *rate*, and *<name of parental cohort>* is available from the patient-flow diagram.

As an example, consider the pseudocode for the **withdraw-transition**. There are two major methods for this object: **pfd-action** and **sm-action**, corresponding to the construction steps performed on the patient-flow diagram and on the statistical model, respectively.

```
(method pfd-action (withdraw-transition,
    source-cohort)
  withdraw-cohort ← create-cohort(yes)
  no-withdraw-cohort ← create-cohort(no)
  connect (source-cohort, withdraw-cohort)
  connect (source-cohort, no-withdraw-
      cohort)
  treatment (withdraw-cohort (source-cohort))
      ← baseline-treatment
)
```

Two new cohorts are created, with differing semantics as regards the withdrawal status of patients in those cohorts. The cohorts are then incorporated into the patient-flow diagram. Finally, an element of domain knowledge is included, that withdrawn patients effectively receive baseline care.

```
(method sm-action(withdraw-transition, source-
    cohort)
  alpha ← create-mixing-parameter(source-
      cohort)
  withdraw-outcome-parameter ← outcome-
      parameter(withdraw-cohort(source-
      cohort))
  no-withdraw-outcome-parameter ←
      outcome-parameter(no-withdraw-cohort
      (source-cohort))
  make-mixture(source-cohort-outcome-
      parameter,
      alpha, withdraw-outcome-parameter,
      no-withdraw-outcome-parameter)
  make-identical(withdraw-outcome-parameter,
      population-baseline-outcome-parameter)
  make-identical(no-withdraw-outcome
      parameter ,population-outcome-
      parameter(treatment(source-cohort))))
```

Here, the mixing model of Fig. 3 is created. The parameters here are all at the sample level; the functions that find the study-level parameters have been eliminated for the sake of simplicity. The action, **make-mixture,** is here performing the main work, converting a parameter (source-cohort-outcome-parameter) formerly identical to a population-level parameter into one that is a deterministic function of other parameters. Domain knowledge in encoded in the instruction to make, the newly created study-level outcome parameters identically deterministic on outcome parameters in the population level,.

## 6  IMPLEMENTATION

The architecture presented here has been implemented in a system called THOMAS (Lehmann 1991), that focuses on problems of internal validity in TAPRCT studies. The influence diagrams are restricted to having one layer of marginally independent chance nodes whose pdfs are beta distributions. All other nodes are deterministic or evidential. Evidence nodes have only one parent. The construction steps are modules whose statistical components are based on the methodological models of Eddy, et al. (Eddy, Hasselblad et al. 1991).

The primary outputs are posterior distributions for the parameters of the model, and expected utility (life expectancy). The algorithm for calculating posterior probabilities uses posterior-mode determination, based on modified Newton–Raphson steepest descent (Shachter 1990), using the approach of Berndt and colleagues (Berndt, Hall et al. 1974). This algorithm requires matrix inversion, an $O(m^3)$ process, where $m$ is the number of parameters with no parents. In our use of this algorithm, parameters that are functionally identical to other parameters are eliminated from the model in $O(n)$ time prior to the posterior-mode determination, where n is the number of statistical parameters. Thus, the overhead of maintaining the influence-diagram levels takes little away from the performance of the algorithm.

Calculation of expected utility is through unidimensional integration.

THOMAS is implemented on a Macintosh II computer. The patient-flow diagram was written in HyperTalk and the remainder of the system was written in Macintosh Common Lisp, using the Common Lisp Object System to house the various objects in the UB architecture. Communication between the two systems used TalkToMe, a precursor to AppleEvents.

A number of different MDSDs and different patient-flow diagrams were tested in this architecture. In each experiment, and as desired, modifications of a major structure did *not* require changes in the other structure.

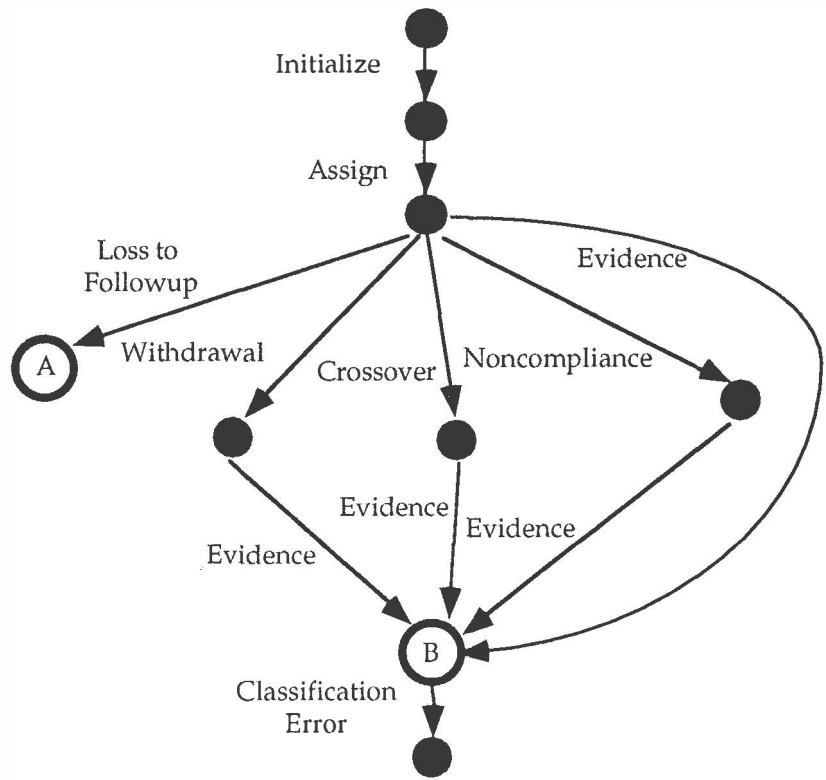

Figure 4: Metadata-state diagram. States represent cohorts, and arcs represent permitted transitions in states of knowledge about cohorts.

A small number of clinicians used THOMAS as intended. The needed tasks were conceptually familiar. Some users disliked the navigational interface, but found the patient-flow diagram to be self-evident (as we expected).

## 7 OTHER WORK

This work reflects on research in the AI-and-statistics community, in the biostatistical community, and in the AI-and-uncertainty community. Statisticians applying AI techniques to statistical problems have, by and large, focused on helping the statistical analyst, rather than the end user (Nelde and Wolstenhome 1986; Lubinsky 1987; Oldford and Peters 1988).

In the biostatistical community, Hasselblad and Eddy have developed the FAST*PRO program (associated with (Eddy, Hasselblad et al. 1991)) to allow medical researchers construct influence diagrams that represent statistical models. Differences with our work include their focus on the data-driven direction of influence-diagram construction and their lack of a semantic interface.

Finally, various researchers of AI and uncertainty have built systems that automatically build influence diagrams. Goldman and Charniak (Goldman and Charniak 1990) have developed a language for constructing influence diagrams, specifically for the domain of natural-language processing. Their system uses a hierarchical and typed influence diagram, as we do; their construction algorithm is more model-driven than is ours. Similarly, Breese (Breese 1987) has built a system that constructs influence diagrams out of modular components. However, he requires the user to have a greater knowledge of influence diagrams than we do.

## 8 CONCLUSION

We have presented an architecture for creating tools that allow end users to interpret the results of scientific studies whose data requires statistical analysis for interpretation. This architecture implements the general decision-analytic approach to scientific inference presented by Lehmann (Lehmann 1990). We believe that this architecture could be extended for other data-analysis problems. For instance, for crossover studies, where patients are first given one, then another treatment, the MDSD and the patient-flow diagram would both have to be modified. Furthermore, the architecture could be extended to account for other methodological concerns. For instance, randomization is a key consideration whose goal is to assure similarity of treatment groups. This concern can be represented at the sample level in terms of relative occurrences of covariate states for study subjects in every treatment arm, and in terms of their relative contributions.

Even more generally, we might imagine that this architecture would be suitable for other AI domains where the primary user (or even knowledge engineer) may wish to be protected from details of Bayesian modeling. Thus, in a vision system, the cognate of the patient-flow diagram might be the hierarchy of visual perception that Levitt and colleagues (Binford, Levitt et al. 1987) have put to such good use, while the MDSD might represent permitted, meaningful sequences of visual processing. Each limb of transition in that process would have implications for the underlying influence diagram as well as for an interface that a user would face.

In each case, there are several tasks for the knowledge engineer. First, he must decide the propriety of the architecture. Second, from observation of users, he must decide on the appropriate semantic metaphor and on the central object of manipulation. Third, from discussions with domain experts, he must encode the rules for transitions. Finally, he must figure out what influence diagram structures and transformations embody the experts' intentions. *This* overall construction task is far from being automated.

In summary, statisticians confront the same problem that AI-system designers do: How should valid inferences be made from real-world data? The user-based architecture can be of service to both these communities.

### Acknowledgments

We thank Bill Brown for his statistical expertise and criticism in our constructing THOMAS, Bill Poland for his help in programming THOMAS, Ruben Kleiman, of Apple Computers for supplying TalkToMe, and Ted Shortliffe, Brad Farr, Holly Jimison, and Eddie Herskovits for constructive discussions in the course of this work. This work was funded by NLM grant LM-07033.